\documentclass[11pt]{article}

\usepackage[final]{acl}

\usepackage{times}
\usepackage{latexsym}

\usepackage[T1]{fontenc}

\usepackage[utf8]{inputenc}

\usepackage{microtype}

\usepackage{inconsolata}

\usepackage{graphicx} 
\usepackage{amsmath} 
\usepackage{amsbsy}  
\usepackage{amsfonts}
\usepackage{times}
\usepackage{helvet}
\usepackage{courier}
\usepackage{xcolor}
\usepackage{siunitx}
\usepackage{booktabs}

\usepackage{graphicx}

\title{
Schema-Adaptive Tabular Representation Learning with LLMs for Generalizable Multimodal Clinical Reasoning}

\author{
  Hongxi Mao\textsuperscript{1,2,$\dagger$} \quad
  Wei Zhou\textsuperscript{1,3,$\dagger$} \quad
  Mengting Jia\textsuperscript{4} \quad
  Tao Fang\textsuperscript{4} \\
  {\bf Huan Gao}\textsuperscript{\textbf{1,5}} \quad
  {\bf Bin Zhang}\textsuperscript{\textbf{4}} \quad
  {\bf Shangyang Li}\textsuperscript{\textbf{1,4*}} \\
  \textsuperscript{1}Beijing University of Posts and Telecommunications, China \quad
  \textsuperscript{2}Boston University, USA \\
  \textsuperscript{3}University of Southern California, USA \quad
  \textsuperscript{4}GDIIST, China \quad
  \textsuperscript{5}Renyixun Health Technology Co., Ltd., China \\
  \texttt{nic\_lab@163.com} \\
}

\newcommand\blfootnote[1]{%
  \begingroup
  \renewcommand\thefootnote{}\footnote{#1}%
  \addtocounter{footnote}{-1}%
  \endgroup
}

\begin{document}
\maketitle

\blfootnote{$^\dagger$Equal contribution.}
\blfootnote{$^*$Corresponding author.}

\begin{abstract}
Machine learning for tabular data remains constrained by poor schema generalization, a challenge rooted in the lack of semantic understanding of structured variables. This challenge is particularly acute in domains like clinical medicine, where electronic health record (EHR) schemas vary significantly.
To solve this problem, we propose Schema-Adaptive Tabular Representation Learning, a novel method that leverages large language models (LLMs) to create transferable tabular embeddings. By transforming structured variables into semantic natural language statements and encoding them with a pretrained LLM, our approach enables zero-shot alignment across unseen schemas without manual feature engineering or retraining. We integrate our encoder into a multimodal framework for dementia diagnosis, combining tabular and MRI data. Experiments on NACC and ADNI datasets demonstrate state-of-the-art performance and successful zero-shot transfer to unseen schemas, significantly outperforming clinical baselines, including board-certified neurologists, in retrospective diagnostic tasks. These results validate our LLM-driven approach as a scalable, robust solution for heterogeneous real-world data, offering a pathway to extend LLM-based reasoning to structured domains. 


\end{abstract}

\section{Introduction}
Machine learning excels in structured data modeling but struggles with cross-domain generalization, particularly in tabular domains where varying schemas cause models to fail across datasets~\cite{topol2019high,rajkomar2019machine,miotto2017deep, lu2025harnessing, zhi2026medgr2}. This limits AI scalability and reliability in fields like healthcare with diverse electronic health records.
At the root of this generalization crisis lies \textit{schema heterogeneity}. Real-world data rarely share consistent column names, coding systems, or data formats. For instance, a key biomarker may appear as a continuous value in one database but as a categorical code in another\cite{jing2026beyond}. Conventional machine learning models are trained on fixed, syntactic representations and lack the ability to reconcile these schema variations, resulting in brittle, non-transferable embeddings \cite{zhang2020harmonization, saeed2020machine, schmid2023challenges, wang2025empowering, shao2025unified, li2025subgraph}. Manual feature harmonization offers a partial remedy but is inherently non-scalable, error-prone, and unsustainable in complex real-world pipelines.

To overcome these limitations, we advocate a new approach from \textit{schema-dependent learning} to \textit{semantic schema understanding}. This work introduces \textbf{Schema-Adaptive Tabular Representation Learning}, a framework that leverages the powerful semantic reasoning of Large Language Models (LLMs) to align heterogeneous structured data through natural language.
Instead of treating column-value pairs as numeric tokens, our approach converts them into semantic text, capturing variable metadata and context. Encoded by a pretrained LLM, these texts produce schema-agnostic embeddings, enabling zero-shot transfer across new datasets without manual alignment or retraining
\cite{narayan2022learning, shin2023tabllm,koloski2025llm, lee2025meme, zheng2026evopi}.

We situate this methodological contribution within a challenging, multimodal setting: differential dementia diagnosis. This task serves as a rigorous \textit{stress test} for our framework, requiring the integration of semantically encoded tabular features with neuroimaging data and the prediction of co-occurring etiologies (e.g., Alzheimer’s and vascular pathologies). By embedding our schema-adaptive encoder into a cross-modal transformer with a multi-objective contrastive learning scheme, we test its robustness under real-world heterogeneity, label imbalance, and limited data availability \cite{gao2022mimic, laura2021multisurv, huang2023gatortron}. Importantly, the medical domain here is not our end goal but a high-stakes validation ground for schema-level generalization—a core capability essential to scalable, language-driven machine learning.

Our contributions are summarized as follows:
\begin{enumerate}
\item We propose a schema-adaptive representation framework that reformulates tabular data as semantically compositional text, allowing pretrained LLMs to achieve zero-shot schema alignment without explicit feature harmonization or fine-tuning.
\item We integrate this encoder into a multimodal architecture to evaluate its robustness under extreme heterogeneity and limited supervision, achieving state-of-the-art performance and outperforming human experts in retrospective diagnostic settings.
\item Through comprehensive analysis, we demonstrate how LLM-based semantic encoding enhances multimodal learning, improves sample efficiency, and yields interpretable decision patterns grounded in domain knowledge.
\end{enumerate}

\section{Related Work}

\subsection{Tabular Representation Learning}
Classical tabular models like Gradient Boosted Decision Trees are purely syntactic and fail to generalize across schemas~\cite{NACCdataset,ADNIdataset}. Recent deep learning approaches, including TabPFN~\cite{TabPFN},TIP~\cite{TIP} and TransTab~\cite{TransTab}, have begun to incorporate semantic information but remain largely schema-dependent, limiting their robustness under significant schema shifts.
The advent of LLMs has catalyzed a new direction. Seminal works in NLP like TaPas~\cite{herzig2020tapas} and other table pretraining methods~\cite{wang2021tuta} have focused on deep, intra-table reasoning for tasks like question answering. Models such as TableGPT~\cite{TableGPT,TableGPT2} and TableDreamer~\cite{zheng2025tabledreamer} extend this to generative tasks, primarily on clean, single-source benchmarks. However, the critical challenge of robust generalization across noisy, heterogeneous real-world schemas remains largely unaddressed.
Our work is positioned to fill this specific gap. We focus on cross-schema generalization, proposing a simple yet robust method that leverages a pretrained LLM as a semantic encoder. By transforming column-value pairs into descriptive statements, our framework achieves semantic alignment across diverse, unseen table structures, a key capability underexplored by prior models focused on single-schema understanding.

\begin{figure*}[t]
    \centering
    \includegraphics[width=0.90\linewidth]{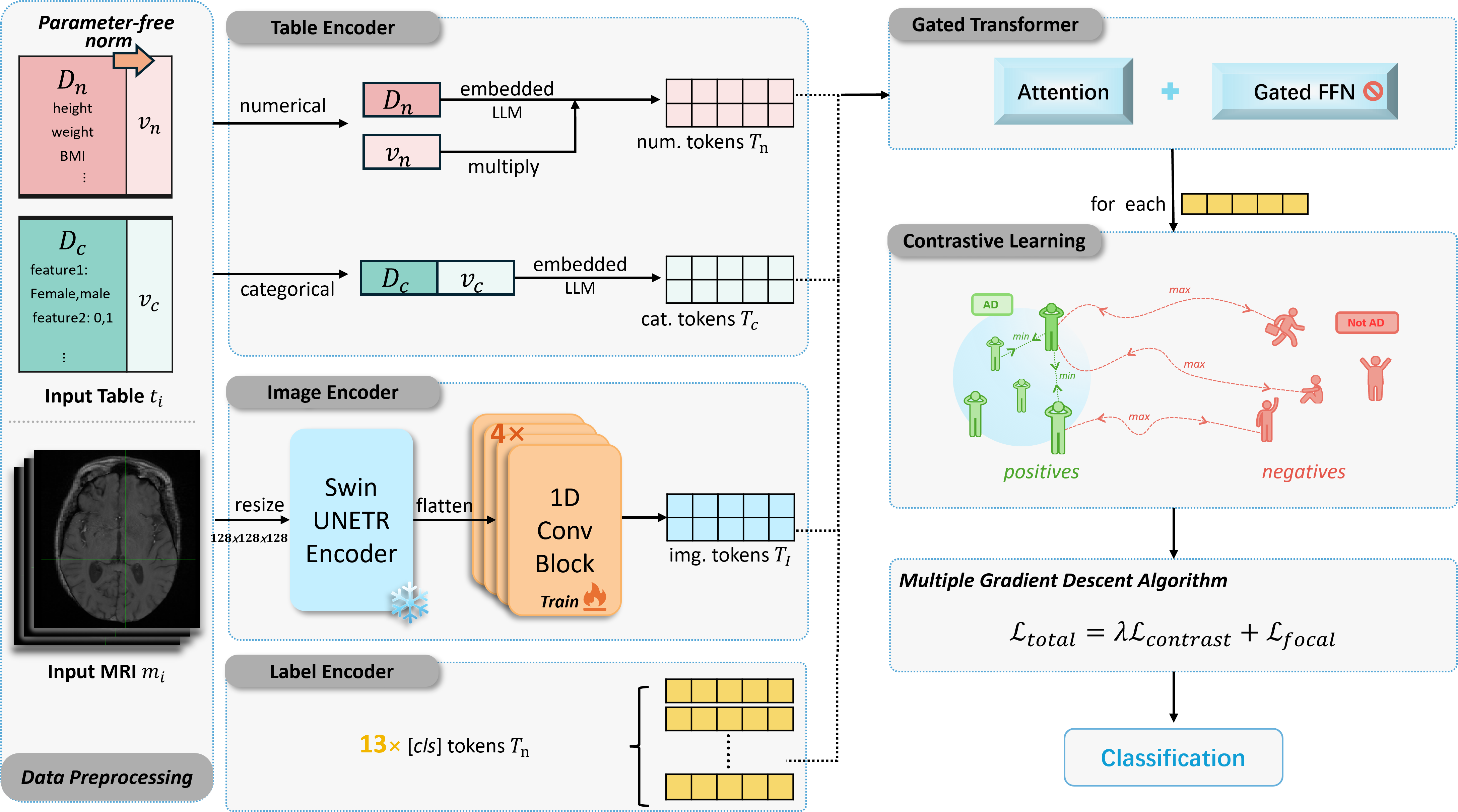} 
    \caption{Overview of our proposed architecture. Patient records comprising tabular and imaging data are processed via modality-specific encoders. The LLM-based table encoder handles schema heterogeneity, while a frozen Swin UNETR encodes images. These representations are fused in a gated transformer with label-specific [CLS] tokens. A composite loss, optimized via MGDA, enables robust multi-label learning under class imbalance.}
    \label{fig:architecture}
\end{figure*}

\subsection{Multimodal Learning with Structured Inputs}


Multimodal learning seeks to unify representations across diverse data types, such as images, text, and structured records, to enable holistic reasoning. Early approaches relied on late-fusion strategies, combining predictions from modality-specific encoders~\cite{survey2024multimodal}, but these often failed to capture cross-modal dependencies during joint representation learning. Recent methods, inspired by CLIP~\cite{radford2021learning}, use contrastive objectives for modality alignment~\cite{hager2023best,hager2023multimodal,TIP}, but image-tabular and text-tabular integration lags due to structured data's lack of semantic continuity. Our framework reinterprets tabular data as linguistic tokens, enabling integration into language-aligned multimodal architectures via cross-modal attention, enhancing performance through language-driven schema representation while ensuring comparability with prior work.

\subsection{Multi-Label Learning and Optimization under Imbalance}
Multi-label classification, prevalent in clinical and biomedical tasks, faces challenges from class imbalance and correlated outputs. Standard supervised contrastive learning (SCL)~\cite{SupCon} performs well for single-label tasks but struggles with co-occurring or hierarchical labels.
Extensions such as MulSupCon~\cite{MulSupCon}, lREG~\cite{lREG}, and JSCL~\cite{JSCL} incorporate label-aware reweighting to address these issues, yet their performance falters on highly imbalanced datasets. Prototype-based methods like C-GMVAE~\cite{GMVAE} and Proto~\cite{PROTO} model label distributions explicitly but are often confounded by high inter-label correlations. 
Our framework uses multi-label classification to evaluate schema generalization, employing a composite objective with focal loss, contrastive regularization, and multi-objective optimization, ensuring performance gains via an LLM-driven, schema-adaptive encoder for robust structured-data learning.

\section{Methodology}

We present a generalizable framework for learning schema-invariant representations from heterogeneous structured data by leveraging the semantic priors of large language models (LLMs). The framework rests on three key components:  
(1) a schema-adaptive tabular encoder that performs language-based semantic tokenization of structured inputs;  
(2) a modality-specific encoder for auxiliary data (e.g., neuroimaging) to assess cross-modal robustness; and  
(3) a unified transformer-based fusion backbone with a multi-objective optimization scheme for complex multi-label prediction.  
Figure~\ref{fig:architecture} illustrates the overall architecture.  
While our evaluation focuses on multimodal dementia diagnosis, the proposed architecture is designed to test a broader hypothesis: \textit{can natural language embeddings serve as a universal representational interface for heterogeneous tabular schemas?}

\subsection{Problem Formulation}

We formalize the task as a multi-label classification problem over multimodal inputs.  
Given a dataset $\mathcal{D} = \{(X_i, Y_i)\}_{i=1}^N$, each instance $X_i$ comprises structured tabular data $x_i^t$ and a complementary modality $x_i^m$ (e.g., an MRI volume).  
The ground-truth label vector $Y_i = [y_{i1}, y_{i2}, \dots, y_{iL}] \in \{0, 1\}^L$ denotes the presence or absence of each condition or class.  
The learning objective is to find a mapping function $f_\theta(X_i)$ that minimizes an aggregate loss over all labels:
\begin{equation}
\min_{\theta} \; \mathbb{E}_{(X, Y) \sim \mathcal{D}} \left[ \ell(f_\theta(X), Y) \right].
\end{equation}
A central requirement is that $f_\theta$ remains robust to arbitrary schema variations in $x_i^t$, enabling zero-shot transfer across unseen datasets without explicit feature harmonization.

\subsection{LLM-Powered Schema-Adaptive Tabular Encoder}

The schema-adaptive encoder serves as the linguistic core of our framework.  
It transforms each column--value pair of a table into a structured textual statement, thereby grounding numerical and categorical features in the semantic space of an LLM.  
Formally, for a tabular input $x_i^t$, we process categorical ($x_c$) and numerical ($x_n$) attributes as follows.

\paragraph{\(x_c \in \text{Categorical Features}\):}  
For a column $c$ with description $D_c$ and value $v_c^i$:  
\begin{enumerate}
    \item \textit{Description Refinement}: A lightweight rewrite function $L$ augments metadata to form human-readable context, e.g., $L(\text{``SEX''}) \rightarrow \text{``Gender of the subject:''}$.  
    \item \textit{Statement Construction}: Construct a natural language statement $S_c^i = L(D_c) \oplus v_c^i$, e.g., ``Gender of the subject: Female''.  
    \item \textit{Semantic Embedding}: Encode $S_c^i$ using a pretrained LLM embedding model to obtain $\boldsymbol{E}_c^i = \mathrm{Emb}(S_c^i) \in \mathbb{R}^d$.  
\end{enumerate}
This process, a form of  \textit{semantic tokenization}, maps schema-specific categorical attributes into a shared language-driven latent space.

\paragraph{\(x_n \in \text{Numerical Features}\):}  
For a column $n$ with description $D_n$ and value $v_n^i$:  
\begin{enumerate}
    \item \textit{Normalization}: Normalize the raw value as $\tilde{v}_n^i = 1 + \frac{v_n^i - \mu_n}{R_n}$, where $\mu_n$ and $R_n$ denote the mean and range of the feature.  
    \item \textit{Description Embedding}: Embed the refined column description using the same LLM encoder: $\boldsymbol{E}_D = \mathrm{Emb}(L(D_n)) \in \mathbb{R}^d$.  
    \item \textit{Value-Weighted Embedding}: Combine magnitude and semantics through element-wise scaling, yielding $\boldsymbol{E}_n^i = \tilde{v}_n^i \cdot \boldsymbol{E}_D \in \mathbb{R}^d$.  
\end{enumerate}
All embeddings $\{\boldsymbol{E}_c^i, \boldsymbol{E}_n^i\}$ are projected through a shared linear layer to dimension $d_{\text{model}} = 256$, forming a unified token sequence $T_{\text{tab}}^i$.  
This encoder effectively decouples model performance from schema syntax, enabling generalization through semantic abstraction.

\subsection{Auxiliary Modality Encoder}

To assess the generality of our schema-adaptive design in multimodal settings, we introduce a secondary encoder for image data.  
Given an MRI volume $x_i^m \in \mathbb{R}^{128 \times 128 \times 128}$ (after standard preprocessing such as skull-stripping and intensity normalization), we employ a frozen Swin UNETR backbone~\cite{SwinU} to extract latent representations $\mathbf{Z} \in \mathbb{R}^{768 \times 4 \times 4 \times 4}$.  
Swin UNETR integrates hierarchical attention and U-Net structures to capture both global and local anatomical context.  
We freeze its pretrained weights to preserve general visual semantics while avoiding overfitting.  
A lightweight projection block comprising four 1D convolutional layers maps $\mathbf{Z}$ into an image token sequence $T_{\text{img}}^i \in \mathbb{R}^{d_{\text{model}}}$.  
Although the imaging modality is domain-specific, it provides a stringent testbed for evaluating the adaptability of language-grounded tabular embeddings in cross-modal reasoning.

\subsection{Multimodal Fusion Backbone}

The token sequences from tabular ($T_{\text{tab}}$) and image ($T_{\text{img}}$) modalities are concatenated along with $L$ learnable \texttt{[CLS]} tokens---one per label---and passed into a gated transformer.  
This \textit{mid-level fusion} design enables deep inter-modality interaction while maintaining representational independence.  
Each \texttt{[CLS]} token $T_k \in \mathbb{R}^{d_{\text{model}}}$, $k \in [1,L]$, acts as a dedicated classifier query.  
After transformer propagation, the contextualized state of each token is used to produce the corresponding label prediction.  
This architecture naturally extends standard text--image transformers (e.g., CLIP) into a text--table--image tri-modal context, where tabular information is expressed in linguistic form.

\subsection{Training Objective for Robust Evaluation}

To manage interdependent labels and severe class imbalance, we formulate training as a multi-objective optimization problem following~\cite{MTLMOO}.  
The total objective comprises $2L$ components---$L$ Focal losses and $L$ contrastive losses---balanced by the Multiple Gradient Descent Algorithm (MGDA).  
MGDA dynamically reweights gradients across objectives, ensuring that high-magnitude tasks do not dominate optimization.  
Formally, we solve:
\begin{equation}
\min_{\alpha^1,\ldots,\alpha^{2L}} 
\left\| \sum_{t=1}^{2L} \alpha^t \nabla_{\theta^{sh}} \mathcal{L}_t \right\|_2^2 
\quad \text{s.t.} \quad 
\begin{aligned}
&\sum_{t=1}^{2L} \alpha^t = 1, \\
&\alpha^t \geq 0.
\end{aligned}
\end{equation}
where $\theta^{sh}$ are shared parameters and $\mathcal{L}_t$ denotes individual loss terms.

\paragraph{Focal Loss.}  
To counteract label imbalance, we adopt the focal loss~\cite{focal}:  
\begin{align}
\ell_{\text{focal}}(p_{i,k}, y_{i,k}) =\ 
& -\alpha_k [ y_{i,k} (1 - p_{i,k})^\gamma \log(p_{i,k}) \notag \\
& + (1 - y_{i,k}) p_{i,k}^\gamma \log(1 - p_{i,k}) ],
\end{align}
where $p_{i,k}$ denotes the predicted probability, $\gamma$ the focusing parameter, and $\alpha_k$ a class-balancing weight.

\paragraph{Multi-Label Contrastive Loss.}  
To further regularize the representation space, we employ a multi-label extension of supervised contrastive learning~\cite{SupCon}, augmented with the hardness-aware dual-temperature mechanism~\cite{dualT}.  
For an anchor representation $r_i^k$ corresponding to label $k$, the contrastive objective is:
\begin{equation}
\begin{split}
\mathcal{L}_{r_i^k} &= -\text{sg}\!\left(\frac{W_\beta^i}{W_\alpha^i}\right) \log \Bigg( \\
&\quad \frac{\sum_{j=1}^N \mathbb{I}_{[y_{jk}=y_{ik}]} \exp\!\left(\frac{r_i^k \cdot r_j^k}{\tau_\alpha}\right)}{\sum_{j=1}^N \exp\!\left(\frac{r_i^k \cdot r_j^k}{\tau_\alpha}\right)} \Bigg),
\end{split}
\end{equation}
where $\text{sg}(\cdot)$ denotes the stop-gradient operator, $\tau_\alpha$ the temperature, and $W_\alpha$, $W_\beta$ are hardness coefficients.  
This contrastive formulation encourages semantic clustering of samples sharing similar label semantics, reinforcing alignment between schema-derived language embeddings and target-level supervision.

\section{Experiments}
We design our experiments to examine how language-grounded representations derived from LLMs enable schema-level generalization and multimodal consistency in structured data modeling. Specifically, we evaluate our framework along three research questions (RQs) that probe complementary aspects of its representational capabilities:
\begin{itemize}
\item \textbf{RQ1: Schema-Level Generalization.} Can the proposed schema-adaptive encoder transfer across entirely unseen tabular schemas without fine-tuning, demonstrating genuine zero-shot representational alignment?
\item \textbf{RQ2: Multimodal Consistency and In-Domain Robustness.} Within a single, complex schema, does linguistic grounding improve in-domain discriminability, and how does the integration of auxiliary modalities (e.g., imaging) contribute to semantic consistency?
\item \textbf{RQ3: Representation Efficiency and Interpretability.} How efficiently can the learned schema-agnostic representations adapt to new domains with limited data, and do the resulting embeddings exhibit interpretable, clinically coherent semantics?
\end{itemize}

\subsection{Experimental Setup}

\paragraph{Datasets for Schema Heterogeneity Evaluation.}
To evaluate schema-level generalization, we use two large-scale dementia datasets with distinct feature schemas. The \textbf{National Alzheimer's Coordinating Center (NACC)} dataset~\cite{NACCdataset} serves as our training and in-domain benchmark, containing over 200{,}000 visit records from 44{,}656 subjects with 390 heterogeneous features and annotations for 12 dementia etiologies (e.g., AD, LBD, VD); a subset provides MRI scans for 308 subjects. The \textbf{Alzheimer's Disease Neuroimaging Initiative (ADNI)} dataset~\cite{ADNIdataset} is reserved solely for zero-shot evaluation, comprising approximately 20{,}000 records from 3{,}392 subjects with a distinct schema of 110 features, and is strictly excluded from all training or validation phases to ensure unbiased unseen-schema testing.

\paragraph{Task Definition.}
We formulate a 12-label dementia etiology prediction task as a high-dimensional, imbalanced, and multimodal classification challenge. Rather than focusing on clinical outcome, this serves as a stress test for schema-level and modality-level generalization under realistic heterogeneity.

\paragraph{Evaluation Metrics.}
We report macro-averaged AUROC as the principal metric, alongside Balanced Accuracy, AUC-PR, and macro F1 Score. These jointly measure discriminative ability, robustness to imbalance, and overall predictive consistency across multiple labels.

\paragraph{Implementation Details.}
Models are trained for 256 epochs using AdamW (lr=0.003) with cosine annealing. The schema-adaptive encoder employs \texttt{text-embedding-3-large} for semantic representations. All experiments run on 4$\times$NVIDIA A40 GPUs under identical computational budgets across baselines to ensure fair comparison. Data was split using official partitions or an 80/10/10 random split stratified by patient ID.

\paragraph{Baselines.}
We benchmark against four representative baselines that capture different perspectives of generalization: (1) \textbf{Human Experts (Neurologists)}---a panel of 12 board-certified neurologists independently assessed 100 multimodal cases. This serves as a real-world clinical benchmark for in-domain diagnostic accuracy. (2) \textbf{TabPFN}~\cite{TabPFN}---a schema-dependent Transformer for tabular data, providing a strong syntactic baseline; (3) \textbf{Gemini-2.5}~\cite{comanici2025gemini}---a large-scale general-purpose multimodal LLM used to evaluate cross-domain reasoning capacity; and (4) \textbf{LLaVA-Med}~\cite{li2023llava}---a domain-adapted medical vision-language model that tests whether specialized finetuning can compensate for schema variance.

\begin{table}[!t]
\centering
\resizebox{\linewidth}{!}{%
\begin{tabular}{l c c c}
\toprule
Model & AD & MCI & Avg-All \\
\midrule
No-LLM (Rand Emb.) & 0.512 & 0.508 & 0.513 \\
No-LLM (Pret Emb.) & 0.625 & 0.611 & 0.611 \\
\textbf{Ours (Schema-Adap.)} & \textbf{0.789} & \textbf{0.765} & \textbf{0.727} \\
\bottomrule
\end{tabular}%
}
\caption{Zero-shot AUROC on unseen ADNI. "Avg-All" = macro-averaged AUROC over 13 diagnostic labels. Our schema-adaptive model outperforms non-semantic baselines.}
\label{tab:zero_shot}
\end{table}

\begin{table*}[!t]
  \centering
    \begin{tabular}{rlrrrrrrr}
    \toprule
          &       & \multicolumn{1}{l}{NC} & \multicolumn{1}{l}{MCI} & \multicolumn{1}{l}{DE} & \multicolumn{1}{l}{AD} & \multicolumn{1}{l}{LBD} & \multicolumn{1}{l}{VD} & \multicolumn{1}{l}{PRD} \\
    \hline
    \multicolumn{1}{l}{Neurologist} 
          & AUROC & 0.930 & 0.699 & 0.914 & 0.761 & 0.833 & 0.613 & 0.517 \\
    \hline
    \multicolumn{1}{l}{\textbf{Ours}} 
          & AUROC & 0.972 & 0.908 & 0.984 & 0.921 & 0.965 & 0.910 & 0.980 \\
    \hline
    \midrule
          &       & \multicolumn{1}{l}{FTD} & \multicolumn{1}{l}{NPH} & \multicolumn{1}{l}{SEF} & \multicolumn{1}{l}{PSY} & \multicolumn{1}{l}{TBI} & \multicolumn{1}{l}{ODE} & \multicolumn{1}{l}{\textbf{Avg}} \\
    \hline
    \multicolumn{1}{l}{Neurologist} 
          & AUROC & 0.708 & 0.719 & 0.517 & 0.613 & 0.497 & 0.516 & 0.680  \\
    \hline
    \multicolumn{1}{l}{\textbf{Ours}} 
          & AUROC & 0.957 & 0.876 & 0.771 & 0.846 & 0.889 & 0.779 & \textbf{0.904}  \\    
    \bottomrule
    \end{tabular}%
      \caption{
     In-domain diagnostic performance on the NACC dataset, measured by AUROC across 13 neurological conditions. Our schema-adaptive model consistently outperforms expert neurologists, especially in less prevalent or complex conditions (e.g., SEF, ODE). ``Avg'' denotes the macro-average AUROC across all 13 labels, reflecting the overall diagnostic robustness of the model.
    }
  \label{tab:vs_neurologist}%
\end{table*}%

\subsection{Results and Analysis}
We now present the empirical results that rigorously evaluate our central hypothesis: language-grounded representations derived from LLMs enable robust schema-level generalization, in-domain multimodal consistency, and interpretable semantic alignment. Results are organized by the three research questions introduced earlier.

\subsubsection{RQ1: Validating Zero-Shot Schema Generalization}
\label{sec:rq1_results}
The first question investigates whether our LLM-powered encoder can bridge the semantic gap between heterogeneous tabular schemas by grounding structured variables in a shared linguistic space. We perform a strict zero-shot cross-dataset evaluation: models are trained exclusively on the NACC dataset and directly tested on the unseen ADNI schema without any fine-tuning or exposure to its feature names or distributions. This setting isolates the model’s intrinsic ability to perform schema-level transfer through semantic abstraction.

We compare our full model against two ablations that explicitly remove linguistic grounding: (1) \textbf{No-LLM (Random Emb.)}, which replaces our encoder with randomly initialized embeddings—representing a purely syntactic baseline dependent on schema-specific structure; and (2) \textbf{No-LLM (Pretrained Emb.)}, which uses a pretrained sentence transformer to embed only raw column names but lacks contextual understanding of feature values. 

As summarized in Table~\ref{tab:zero_shot}, both non-LLM baselines collapse when evaluated on the unseen ADNI schema, with performance barely above random, confirming their brittleness to schema shifts. In contrast, our schema-adaptive model achieves an average AUROC of 0.727, a substantial margin demonstrating its capacity to align heterogeneous feature semantics. For instance, the model successfully maps \textit{``MMSE Total Score''} from NACC and \textit{``MMSCORE''} from ADNI into a coherent semantic embedding. This finding provides strong empirical evidence that linguistic tokenization allows the model to perform cross-schema translation, enabling zero-shot reasoning across disparate tabular structures—a core capability long considered unattainable for standard tabular models.

\subsubsection{RQ2: In-Domain Performance and Multimodal Consistency}
Having established zero-shot transfer, we next evaluate whether the proposed schema-adaptive encoder maintains strong discriminative performance within a single domain and how linguistic grounding influences multimodal representation quality. The evaluation uses the NACC dataset as an in-domain benchmark, comparing our model against human experts and state-of-the-art AI baselines.

\paragraph{Comparison with Human Experts.}
To anchor performance in a real-world context, we benchmark our model against the diagnostic decisions of 12 board-certified neurologists. As shown in Table~\ref{tab:vs_neurologist}, our model achieves a macro-averaged AUROC of 0.904, representing a strong 32.9\% relative improvement over the human experts’ average of 0.680. The advantage is particularly pronounced in complex etiologies with ambiguous symptoms—e.g., Systemic and Endocrine Factors (SEF), where the model (AUROC 0.771) significantly exceeds the expert baseline (0.517). This demonstrates that language-grounded embeddings enable integrative reasoning over subtle, high-dimensional patterns that are difficult for human clinicians to aggregate consistently.

\begin{table}[!t]
  \centering
  \resizebox{\linewidth}{!}{%
    \begin{tabular}{lcccc}
    \toprule
    Model & AUROC & Bal Acc & AUC(PR) & F1 \\
    \midrule
    TablePFN       & 0.868          & 0.683            & 0.505            & 0.439       \\
    Gemini-2.5     & -              & 0.663            & -                & -           \\
    LLaVA-Med      & -              & 0.589            & -                & -           \\
    \midrule
    \textbf{Ours}  & \textbf{0.904} & \textbf{0.785}   & \textbf{0.533}   & \textbf{0.476} \\
    \bottomrule
    \end{tabular}%
  }
    \caption{In-domain performance (NACC dataset). AUROC, Bal Acc, AUC(PR), and F1 are evaluated. TablePFN/ours output probabilities (full metrics); Gemini-2.5/LLaVA-Med only output binaries (- = not applicable).}
  \label{tab:vs_ai_baselines}%
  \vspace{-1mm}
\end{table}%

\paragraph{Comparison with AI Baselines.}
Table~\ref{tab:vs_ai_baselines} compares our framework to state-of-the-art tabular and multimodal models. Our schema-adaptive architecture consistently outperforms all baselines, achieving 0.904 AUROC versus 0.868 for TablePFN, confirming that linguistic grounding not only supports schema transfer but also improves in-domain representation quality. The model’s Balanced Accuracy (0.785) also exceeds large-scale general-purpose multimodal LLMs such as Gemini-2.5 (0.663) and LLaVA-Med (0.589), illustrating that scale alone cannot compensate for semantic misalignment in structured data. Together, these results show that our approach achieves both precision and interpretive stability within complex multimodal settings, establishing a new standard for schema-adaptive tabular reasoning.

\subsubsection{RQ3: Representation Efficiency and Interpretability}
\paragraph{Low-Resource Adaptation and Sample Efficiency.}
An essential theoretical advantage of schema-grounded representations is efficient adaptation under data scarcity. We fine-tune the NACC-pretrained encoder on progressively smaller subsets of the ADNI dataset to assess how well linguistic priors accelerate convergence and transfer. As shown in Table~\ref{tab:fewshot_results}, even with only 300 samples, our fine-tuned model achieves an AUROC of 0.9362, surpassing both a model trained from scratch on ADNI (0.7206) and even one trained on the full dataset (0.8943). This indicates that schema-level linguistic alignment transfers semantic knowledge compactly, allowing the model to achieve near-optimal performance with minimal labeled data—a highly desirable property for medical and scientific domains with limited annotations.

\begin{table}[ht]
\centering
\resizebox{\linewidth}{!}{
\begin{tabular}{l S[table-format=1.4] S[table-format=1.4]}
\toprule
{Train Samples} & {NACC-FewShot} & {ADNI-Train} \\
\midrule
30    & \textbf{0.7389} & 0.6982 \\
100   & \textbf{0.7561} & 0.7176 \\
300   & \textbf{0.9362} & 0.7206 \\
1000  & \textbf{0.9520} & 0.8943 \\
2713  & \textbf{0.9532} & 0.9320 \\
\bottomrule
\end{tabular}}
\caption{Few-shot generalization (AUROC) on ADNI dataset. NACC-FewShot: fine-tuned schema-grounded model; ADNI-Train: model trained from scratch.}
\label{tab:fewshot_results}
\end{table}

\begin{figure}[!t]
    \centering
\includegraphics[width=0.94\linewidth]{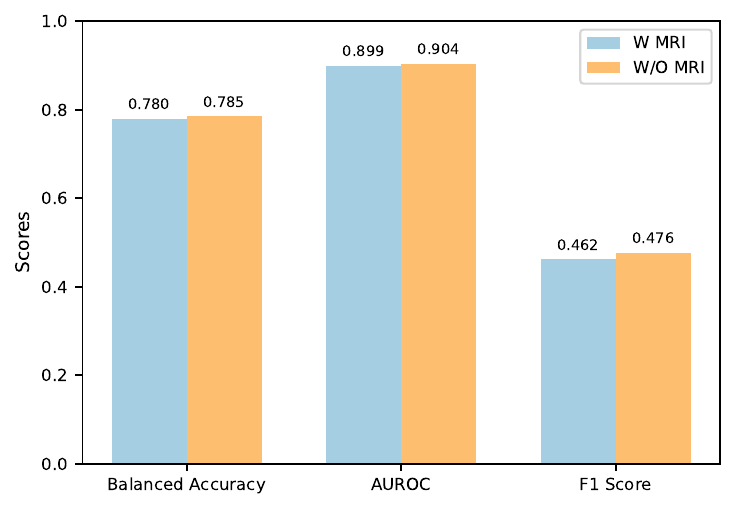}
    \caption{Comparison between table-only and multimodal variants.}
    \label{fig:mri+t_vs_t}
\end{figure}

\begin{figure}[!t]
    \centering
    \includegraphics[width=0.94\linewidth]{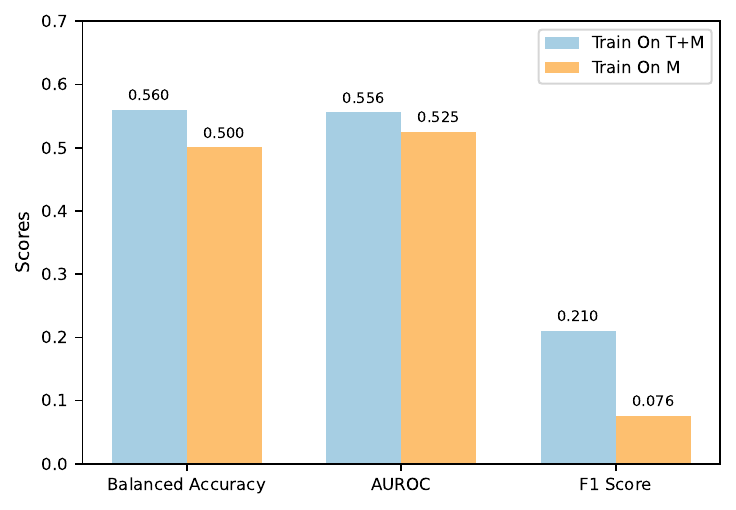}
    \caption{Image-only vs. multimodal training performance.}
    \label{fig:mri vs t+m}
    \vspace{-4mm}
\end{figure}

\paragraph{In-Depth Analysis: The Role of Multimodality.}
A nuanced trend emerges in multimodal integration. While adding MRI data yields only marginal overall improvement (Figure~\ref{fig:mri+t_vs_t}), it plays a critical regularization role. As shown in Figure~\ref{fig:mri vs t+m}, models trained on MRI alone overfit severely, whereas co-training with semantically grounded tabular representations constrains the image encoder to align with language-informed feature semantics. This demonstrates that linguistic priors act as a stabilizing anchor for weak or data-scarce modalities, reinforcing the hypothesis that language-based abstraction can unify heterogeneous signals within a coherent representational manifold.

\paragraph{Architectural Ablations.}
We further examine the sensitivity of architectural design choices. Table~\ref{tab:layer_ablation} shows that a 2-layer Transformer achieves the best trade-off between model capacity and generalization, outperforming both shallower and deeper variants. Similarly, Table~\ref{tab:projector_ablation} reveals that a single linear projection outperforms a deeper MLP, suggesting that our LLM-based encoder already provides a structured and semantically meaningful latent space. These findings confirm that the observed performance gains originate from the language-driven representation itself rather than architectural complexity.

\begin{table}[!t]
\centering
\renewcommand{\arraystretch}{1.2}
\small
\setlength{\tabcolsep}{8pt}
\resizebox{\linewidth}{!}{%
\begin{tabular}{lccc}
\hline
Metric & 1 Layer & 2 Layers & 3 Layers \\
\hline
F1 Score & 0.4170 & \textbf{0.4757} & 0.3959 \\
Bal Acc & 0.7423 & \textbf{0.7853} & 0.7072 \\
AUROC & 0.8672 & \textbf{0.9044} & 0.8537 \\
AUC(PR) & 0.4793 & \textbf{0.5334} & 0.4553 \\
\hline
\end{tabular}%
}
\caption{Performance across transformer depths.}
\label{tab:layer_ablation}
\smallskip
\end{table}

\begin{table}[!t]
\centering
\renewcommand{\arraystretch}{1.2}
\small
\setlength{\tabcolsep}{10pt}
\resizebox{\linewidth}{!}{%
\begin{tabular}{lcc}
\hline
Metric & Single Linear & 2-Layer MLP \\
\hline
F1 Score & \textbf{0.4757} & 0.4133 \\
Bal Acc & \textbf{0.7853} & 0.7435 \\
AUROC & \textbf{0.9044} & 0.8691 \\
AUC(PR) & \textbf{0.5334} & 0.4799 \\
\hline
\end{tabular}%
}
\caption{Projection architecture ablation.}
\label{tab:projector_ablation}
\end{table}

\subsubsection{Interpretability: What Does the Model Learn?}
\begin{figure}[!t]
    \centering
\includegraphics[width=0.99\linewidth]{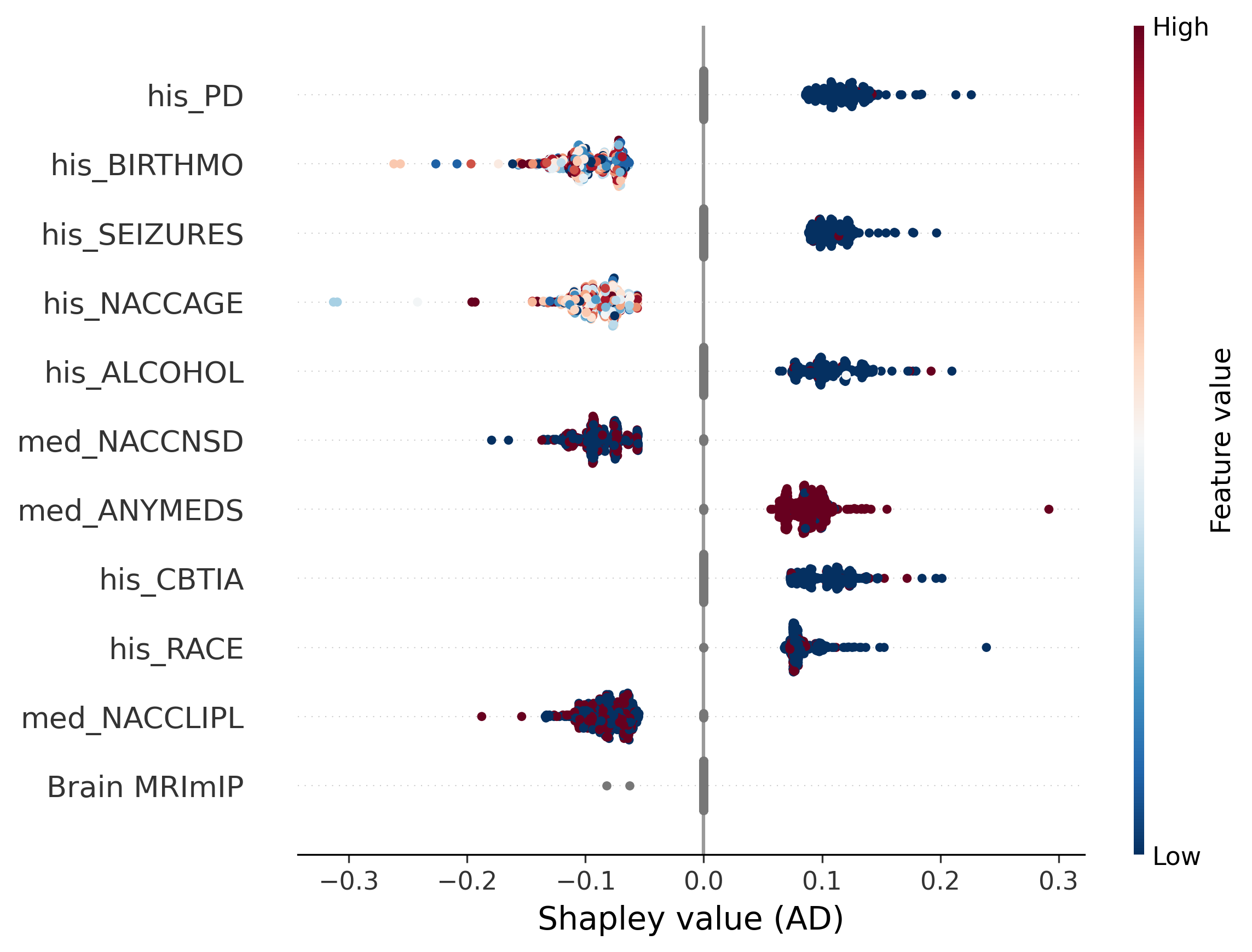}
    \caption{SHAP summary plot for AD prediction showing the top 10 features and an MRI-derived feature. Each point represents a patient; the x-axis indicates Shapley value impact and color denotes feature value (blue=low, red=high).}
    \label{fig:shap_mci}
    \vspace{-1mm}
\end{figure}

Finally, to verify that the learned embeddings reflect meaningful, human-interpretable semantics, we employ SHAP~\cite{shap} analysis for Alzheimer’s Disease (AD) prediction. Figure~\ref{fig:shap_mci} shows that the model highlights clinically coherent predictors such as seizure history (\texttt{his\_SEIZURES}) as positive indicators and Parkinson’s history (\texttt{his\_PD}) as a negative predictor. Notably, MRI-derived features are absent from the top ten contributors, whereas aggregated medical and medication histories dominate. This indicates that the model’s decision process arises from linguistically grounded clinical reasoning rather than superficial correlations. In essence, our schema-adaptive LLM encoder learns to represent structured variables in a semantically interpretable manner, aligning with established domain knowledge and demonstrating the emergence of genuine language-based understanding.

\section{Conclusion}
In this work, we addressed the challenge of schema heterogeneity in tabular data by proposing Schema-Adaptive Tabular Representation Learning, which leverages large language models to recast structured variables as semantically grounded natural language statements. By encoding column descriptions and values as language-based tokens, our method decouples model behavior from brittle syntactic formats, enabling robust zero-shot alignment across heterogeneous schemas. We validated the framework through multimodal dementia diagnosis as a rigorous testbed: the model achieves strong transfer to unseen schemas, sets new in-domain benchmarks, and demonstrates marked sample efficiency under limited-data adaptation. These results support our central claim that natural language can serve as a powerful interface for heterogeneous structured data, guiding the development of more generalizable and trustworthy AI systems. 
Beyond this domain, the framework offers a scalable foundation for schema-agnostic learning across broader multimodal and structured reasoning tasks.

\section{Acknowledgments}
This work was supported in part by the Young Scientists Fund of the National Natural Science Foundation of China (Grant No.~32500997, S.~Li), and in part by Beijing Renyixun Health Technology Co., Ltd.

\section*{Limitations}
The effectiveness of our framework is defined by several key design choices, which also delineate important areas for future work.

First, the performance of our semantic encoder is contingent on the availability of reasonably descriptive metadata (i.e., column names). In "low-context" scenarios with cryptic or absent feature names (e.g., "Var1"), the model's performance may gracefully degrade towards that of syntax-dependent baselines, as the LLM has limited semantic signal to leverage. This highlights a boundary condition of our approach and suggests a promising research direction in automatically inferring semantics for poorly annotated tabular data.

Second, our implementation relies on a specific pretrained LLM embedding model (OpenAI's text-embedding-3-large). While our results demonstrate the viability of this approach, the representational quality is naturally tied to the capabilities of the chosen foundational model. A comprehensive, comparative study of different open-source and proprietary LLMs, as well as an analysis of their potential downstream bias propagation, was beyond the scope of this work but remains a critical step for developing production-ready systems.

Third, our empirical validation was deliberately situated in a complex and high-stakes medical domain to serve as a rigorous testbed. While this provides strong evidence of the framework's robustness, its performance characteristics on tabular data from other domains (e.g., finance, e-commerce) have not yet been evaluated. Establishing the framework's broader, domain-agnostic applicability constitutes an important and exciting next step, building upon the foundational evidence presented in this paper.

\bibliography{custom}

\appendix



\end{document}